\documentclass[10pt,twocolumn,letterpaper]{article} 

\usepackage{avss}
\usepackage{times}
\usepackage{epsfig}
\usepackage{graphicx}
\usepackage{amsmath}
\usepackage{amssymb}
\usepackage{diagbox}
\usepackage{caption}
\usepackage{diagbox}
\usepackage{authblk}
\usepackage{subcaption}
\usepackage{adjustbox}

\usepackage[pagebackref=true,breaklinks=true,colorlinks,bookmarks=false]{hyperref}

\avssfinalcopy 
\ifavssfinal\fi 
\def\avssPaperID{50} 

\begin{document}

\title{FLAME: Facial Landmark Heatmap Activated Multimodal Gaze Estimation}
\author[1,2]{Neelabh Sinha}
\author[1,3]{Michal Balazia}
\author[1,3]{François Bremond}
\affil[1]{INRIA Sophia Antipolis - M\'{e}diterran\'{e}e, France}
\affil[2]{Birla Institute of Technology and Science, Pilani, India}
\affil[3]{Université Côte d'Azur, France}
\affil[ ]{\href{https://github.com/neelabhsinha/flame}{https://github.com/neelabhsinha/flame}}
\maketitle
\ifavssfinal\thispagestyle{empty}\fi 

\begin{abstract}
3D gaze estimation is about predicting the line of sight of a person in 3D space. Person-independent models for the same lack precision due to anatomical differences of subjects, whereas person-specific calibrated techniques add strict constraints on scalability. To overcome these issues, we propose a~novel technique, Facial Landmark Heatmap Activated Multimodal Gaze Estimation (FLAME), as a~way of combining eye anatomical information using eye landmark heatmaps to obtain precise gaze estimation without any person-specific calibration. Our evaluation demonstrates a~competitive performance of about 10\% improvement on benchmark datasets ColumbiaGaze and EYEDIAP. We also conduct an~ablation study to validate our method.
\end{abstract}

\section{Introduction}

Eye gaze is an important non-verbal cue of humans and is capable of describing human emotion and behaviour. It is used in large cross-domain applications such as study of human behavior~\cite{ab}, clinical diagnosis~\cite{ad}, human-computer and human-robot interaction~\cite{ag,af}. To have precision in such applications, it is very important to have an accurate estimation of eye gaze. 3D gaze estimation is about inferring the actual line-of-sight of a person in 3D space and is the main focus of this paper.

\begin{figure}
\centering
\includegraphics[width=\linewidth]{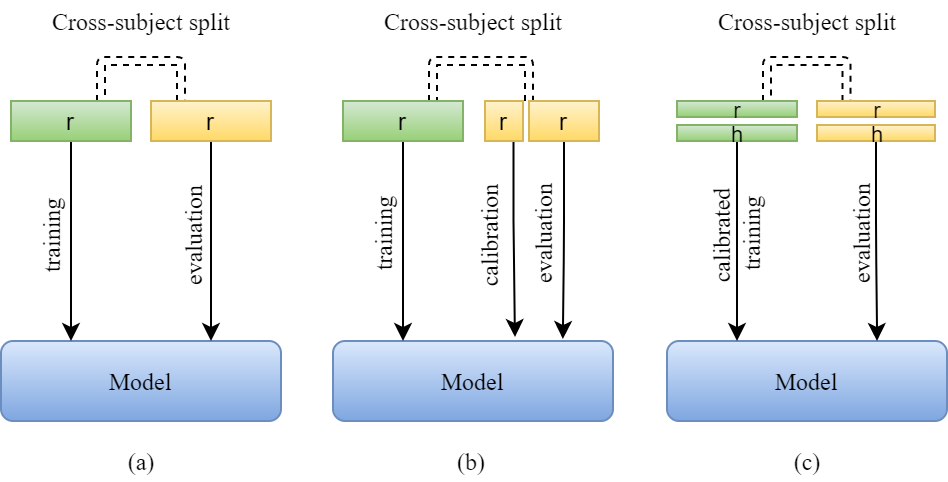}
\caption{Different types of gaze estimation methods: (a)~person-independent technique, (b)~person-specific technique, (c)~\textbf{FLAME}. Training subjects are green and test subjects are yellow. $r$ stands for RGB image and $h$ stands for eye landmark heatmap.}
\label{fig:different_methods}
\end{figure}

Owing to the importance of study of this problem, vast research has been conducted. With the evolution in the field of deep learning, this problem has also moved significantly towards using it to predict gaze~\cite{ak,ao,am,an}. Many techniques focus on person-independent gaze estimation~\cite{am,an}~(Figure~\ref{fig:different_methods}(a)) where the model is evaluated on un-encountered subjects. But gradually, it was proved that it is not possible to reduce error below a certain point due to heavy dependency of gaze on the anatomy of the eye of different subjects~\cite{ap}. Few classical model-based techniques~\cite{ap,ar} were able to incorporate these anatomical features with few calibration samples, but they are not robust to variation in the image under different settings. As a result, subject-specific personalized models~\cite{ao,aq}~(Figure~\ref{fig:different_methods}(b)) got introduced which were able to further reduce errors using additional calibration steps on the test subjects to incorporate anatomical features. 

However, although these person-specific models tend to give a low value of error, using them to design practical systems is difficult. This is because, in real-world scenarios, systems keep on encountering new subjects. So, for additional calibration, there have to be steps to take test samples with accurate ground truth and feed it to the model to calibrate before the actual task can be carried out. In different settings with varying surroundings and distance between subject and camera, it is challenging to procure test samples with accurate ground truth to calibrate the model. Further, for applications like clinical diagnosis of psychological disorders, for example autism and schizophrenia, where gaze is a very important non-verbal cue, the affected people cannot follow precise instructions to be performed for acquiring calibration samples. This makes it practically impossible to rely on person-specific methods for many applications.

In order to address this issue, and also use the anatomical features while predicting the gaze, we propose the Facial Landmark Heatmap Activated Multimodal Gaze Estimation (FLAME), which focuses on precise gaze estimation without calibration by incorporating the anatomical information contained in eye landmarks along with RGB features~(Figure~\ref{fig:different_methods}(c)). Landmarks corresponding to the outline of the eye hold key information about its anatomical features, and as also shown by~\cite{bc}, depict strong correlation with gaze direction. However, since the extraction of eye landmarks using other algorithms are also prone to errors, to make the model robust to these, we take a heatmap based approach inspired by~\cite{as} for skeleton-based action recognition. Instead of using absolute coordinates of landmark points, we use a Gaussian probability distribution heatmap for each landmark. This allows us not to pose strong constraints on the obtained value of the landmark coordinates and rather to provide it in the form of a probability distribution, while also allowing the system to extract salient anatomical features from it.

\begin{figure}[htbp]
\centering
\includegraphics[width=\linewidth]{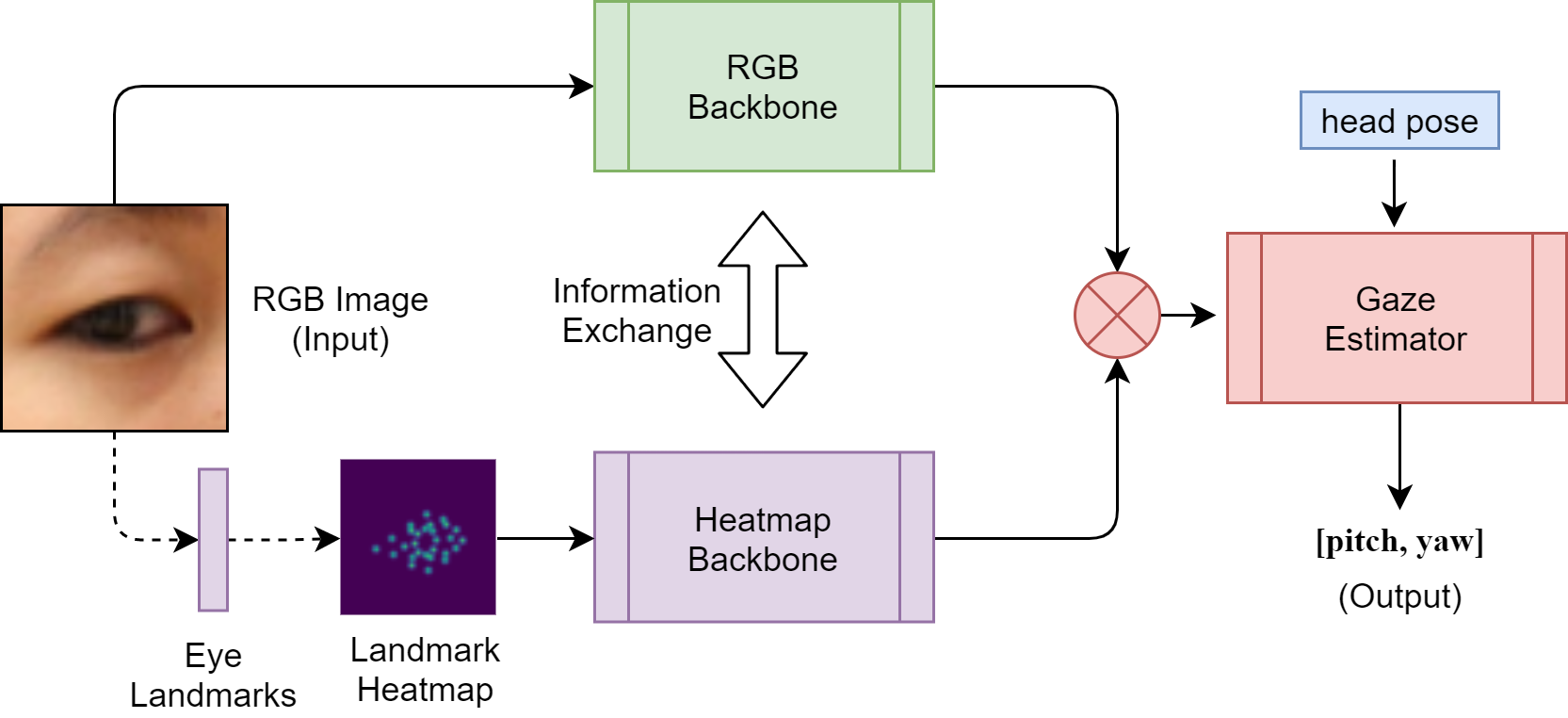}
\caption{Method Summary. Landmark heatmap is first obtained from cropped eye images. Then, both are processed through a two-stream CNN-based network to extract gaze and anatomical information separately, which is further used along with head pose to predict gaze. Dotted line marks the pre-training extraction and solid line marks the training pipeline.}
\label{fig:model_block}
\end{figure}

The high-level idea is depicted in Figure~\ref{fig:model_block}, which follows a two-stream CNN based neural network, working on RGB and eye landmark heatmap modalities to predict gaze angles. These angles can easily be converted to directional vector using basic trigonometry. To the best of our knowledge, this is the first work in gaze estimation using the information from eye landmarks in the form of a heatmap along with the RGB image.

Key contributions of this work are as follows:
\begin{itemize}
\item A novel approach for \textbf{precise gaze estimation without calibration by incorporating anatomical features} of the eye seamlessly with only RGB image as input.
\item A method for obtaining, and \textbf{using eye landmarks as heatmaps} to extract anatomical features and merge with RGB feature map.
\item Extensive experimentation on public benchmarks to obtain \textbf{accurate gaze estimations in person-independent settings}.
\end{itemize}

In the rest of the paper, we discuss our proposed approach in detail. Section~\ref{sec-rel} describes the past work in this field, followed by complete details of the proposed methodology in Section~\ref{sec-meth}. In Section~\ref{sec-impl} we discuss the implementation details, which is succeeded by results and ablations in Section~\ref{sec-res}. Section~\ref{sec-con} concludes the work and discusses its future scope.

\section{Related Work}
\label{sec-rel}

3D gaze estimation can be broadly divided into 2 parts: geometric methods~\cite{ar} and appearance-based methods~\cite{av}. Geometric methods~\cite{ap,ar} focus on modelling the eye, extracting features in terms of geometrical parameters and use it to predict gaze. They have several constraints like near-frontal head-pose and high-resolution images, and thus have limited application as compared to appearance-based methods, which can directly estimate gaze from eye images, and thus have gained significant popularity in present days.

Among person-independent methods,~\cite{am} proposed a LeNet-based architecture using a single eye patch,~\cite{ak} used parallel VGG-16 based architecture and fed eyes and face to estimate head pose and gaze angle,~\cite{ca} used Dilated Convolutions, while~\cite{an} used a more complex CNN backbone to predict gaze.~\cite{ba} did full-face gaze estimation by generating an attention map, and~\cite{bb} implemented three parallel CNN backbones to process left eye, right eye and full face to predict gaze.~\cite{be} depicted a more robust use of head-pose for gaze estimation. Further,~\cite{bc} proposed a multi-task CNN to predict eye-gaze and facial landmarks, and concluded that eye-landmarks are strongly co-related to gaze. As a developing trend, to remove person-specific bias of the models as shown by~\cite{ap}, person-specific models~\cite{ao,aq} have been proposed for gaze estimation which calibrate the model on few samples of test subjects before evaluation. Unsupervised learning has also come into the picture with~\cite{bl} proposing unsupervised gaze estimation model using gaze redirection mechanism. Recently,~\cite{bm} proposed gaze estimation using transformers and~\cite{bn} implemented gaze estimation by attention mechanism and also using difference layer to remove unwanted features from both eyes.

Multimodal fusion has been of significant interest across domains. With time, late fusion has gained significant popularity where each mode is processed separately and combined later. It can be achieved by element-wise summation, concatenation, bi-linear product~\cite{bp}, weighted average~\cite{bo}, rank minimization~\cite{bq}.~\cite{br} involves using attention to pick the best mode for each input. Multimodal fusion module~\cite{au} was proposed which allows slow modality fusion in features of different dimensions.

Inspired by these works, we developed our own technique which we describe in-depth in the following sections.

\section{Method Overview}
\label{sec-meth}

The first part of our method is to extract the eye landmarks from the RGB image and generate the heatmap, followed by a two stream network having multiple components. From here on, for all discussions, $r$ refers to RGB image and $h$ represents eye landmark heatmap.

\subsection{Eye Landmark Heatmap}
\label{sec:heatmap_generation}

\begin{figure}[htbp]
    \centering
    \includegraphics[width=\linewidth]{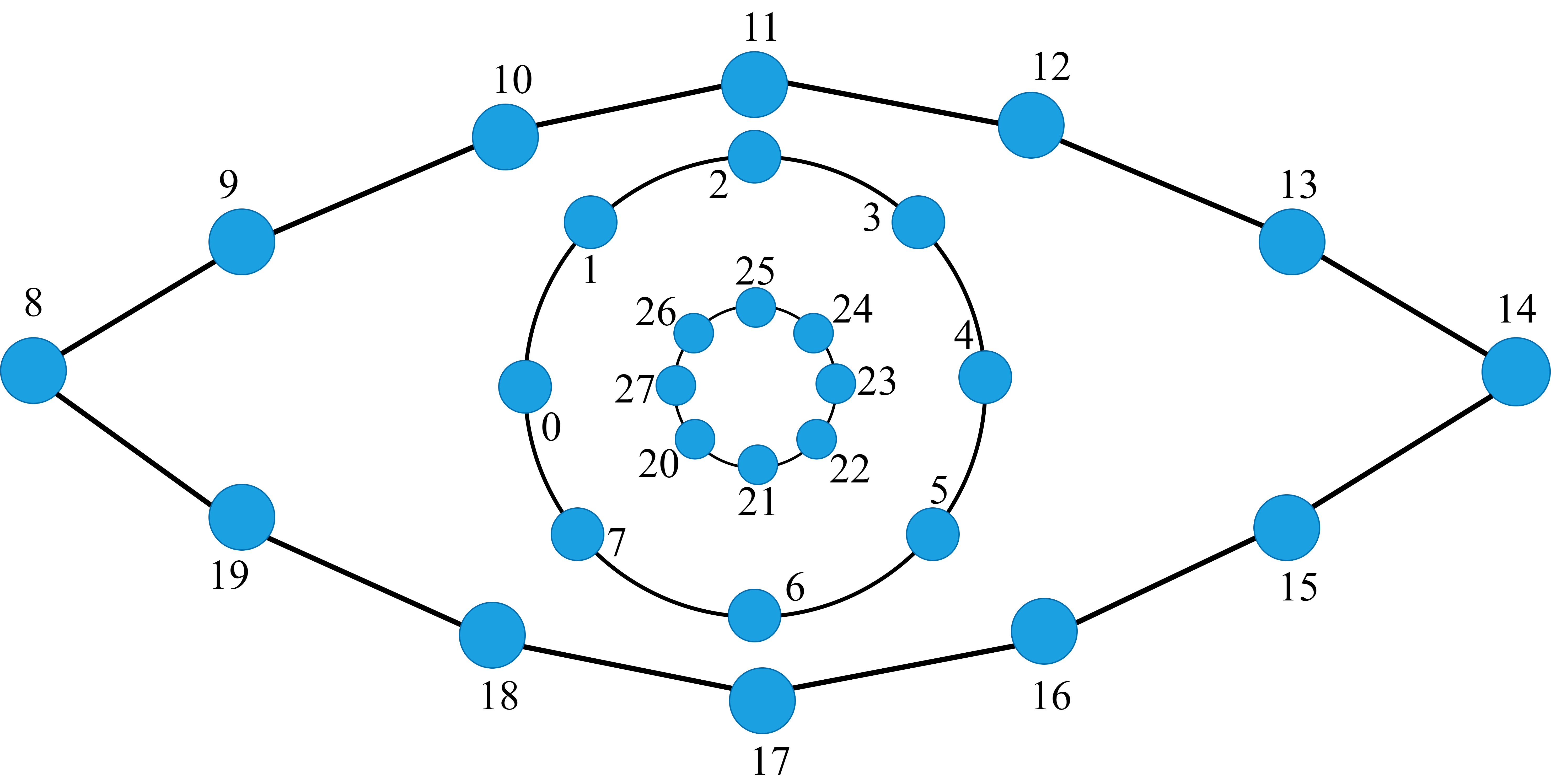}
    \caption{Eye landmarks from Openface 2.0}
    \label{fig:eye_landmarks}
\end{figure}

To get the heatmap, we first extract the 2D eye landmarks from corresponding RGB images using facial landmark submodule~\cite{bv,bu} of OpenFace~2.0~\cite{bt}. We prefer this because it can provide landmarks for the outline of the cornea and pupil along with outer eye, and is significantly precise and easy to use. OpenFace gives 28 2D landmarks for each eye as shown in Figure \ref{fig:eye_landmarks} in pixel coordinates, capturing the outline of the complete eye, cornea and pupil. These three components not only allow us to capture the shape of the outer eye but also the boundaries of inner regions, giving a large variety of descriptive information about the anatomy of the eye.

For a $d$-dimensional input, Gaussian probability distribution heatmap is defined by
\begin{equation}
P(X) = \frac{1}{\sqrt{\left(2\pi\right)^d\left|{\Sigma}\right|}}\exp\!\left({-\frac{1}{2}\left(\left(X\!-\!\mu\right)^\intercal\Sigma^{-1}\left(X\!-\!\mu\right)\right)}\right)
\label{eqn:gaussian_pdf}
\end{equation}
with mean $\mu$ and covariance matrix $\Sigma$. We use it to obtain the Gaussian probability distribution heatmap of the eye landmark coordinates having the same dimensions as the input image.

\subsection{Transfer Function}

\begin{figure}[htbp]
\centering
\includegraphics[width=\linewidth]{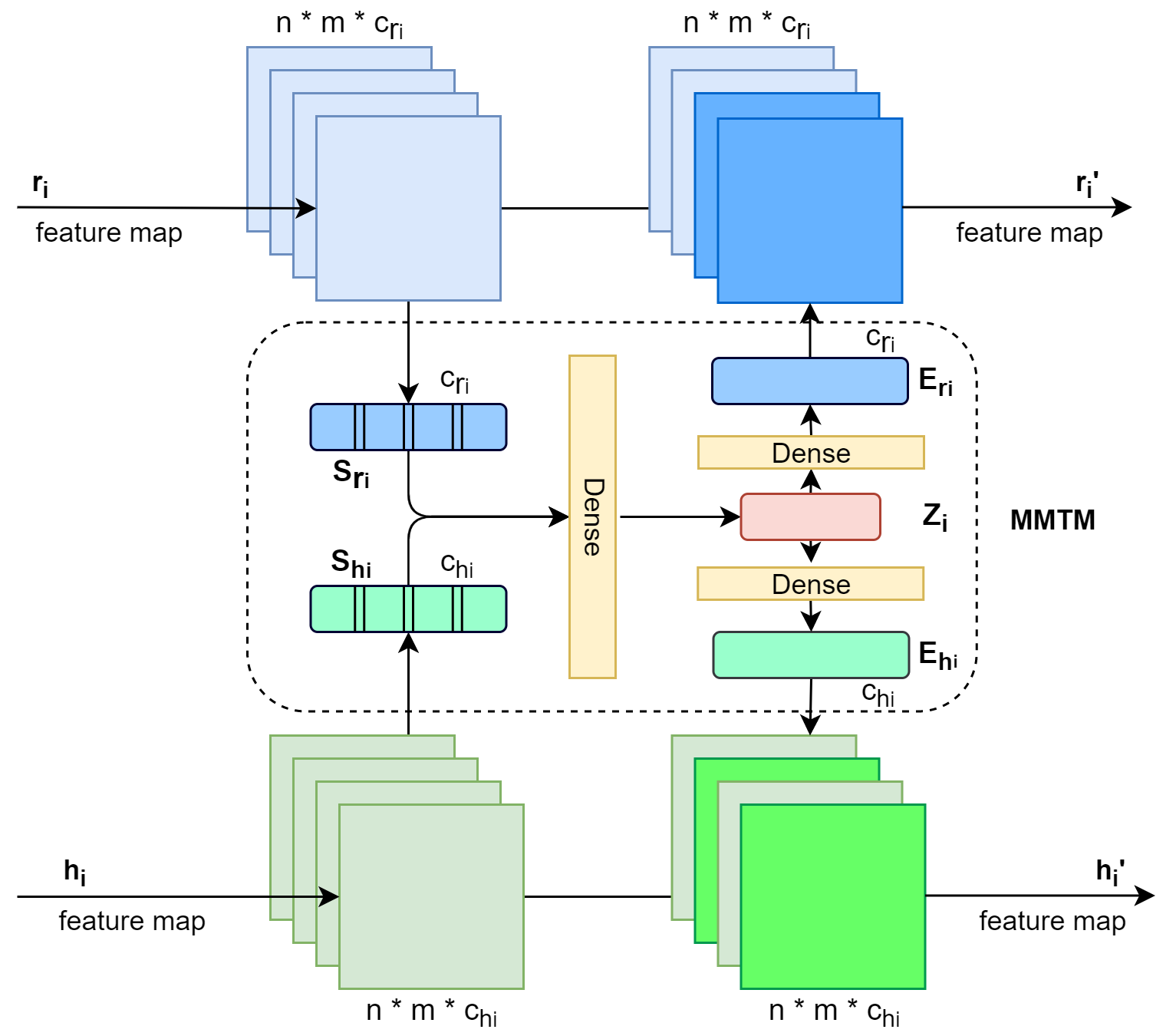}
\caption{Multimodal Transfer Module~\cite{au}}
\label{fig:mmtm}
\end{figure}

The transfer function $T\left(r,h\right)$ is the function that defines the exchange of information between the two streams. It is based on Multimodal Transfer Module~(MMTM)~\cite{au}, which is a slow modality fusion block used to re-calibrate channel-wise features based on squeeze and excitation mechanism between any two feature maps of arbitrary dimension. The complete set of steps can be visualized from Figure~\ref{fig:mmtm}.

\begin{figure*}
\includegraphics[width=\textwidth]{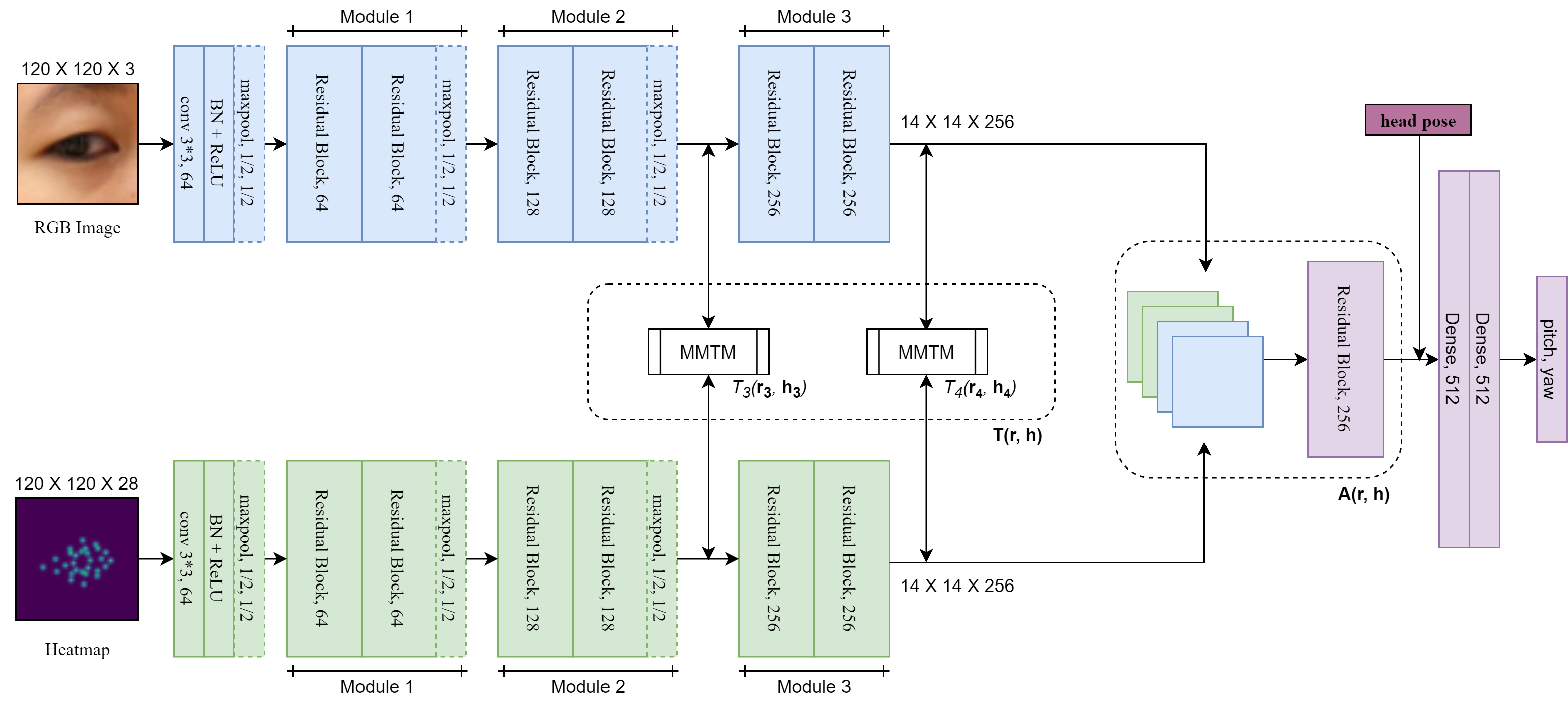}
\caption{Complete model architecture. Two identical CNN backbones process the RGB and heatmap modality separately, with exchange of information at intermediate layers as per transfer function and later fusing them using aggregation function. This hybrid feature map along with head pose is passed to a fully-connected regression network to predict gaze angles.}
\label{fig:complete_model}
\end{figure*}

Assume an input feature vector of height $n$ and width $m$. Let $\mathbf{r_i}\in\mathbb{R}^{n \times m \times c_{r_{i}}}$ be the $i$-th RGB feature map of $c_{r_{i}}$ channels and $\mathbf{h_i}\in\mathbb{R}^{n \times m \times c_{h_{i}}}$ the $i$-th heatmap feature vector of $c_{h_{i}}$ channels. We first squeeze them into $\mathbf{S_{r_i}}\in\mathbb{R}^{c_{r_i}}$ and $\mathbf{S_{h_i}}\in\mathbb{R}^{c_{h_i}}$, the channel-wise descriptors, using global average pooling across the spatial dimensions:
\begin{equation}
S_{r_i}\left(c\right) = \frac{1}{n \cdot m}\sum_{p=1}^{n} \sum_{q=1}^{m}r_i\left(p,q,c\right)
\label{eqn:mmtm_squeeze_r}
\end{equation}
\begin{equation}
S_{h_i}\left(c\right) = \frac{1}{n \cdot m}\sum_{p=1}^{n} \sum_{q=1}^{m}h_i\left(p,q,c\right)
\label{eqn:mmtm_squeeze_h}
\end{equation}
Then, we concatenate these two, and pass it through a dense layer to obtain a joint representation $\mathbf{Z_i}$. This $\mathbf{Z_i}$ is passed through separate dense layers to obtain excitation signals $\mathbf{E_{r_i}}\in\mathbb{R}^{c_{r_i}}$ and $\mathbf{E_{h_i}}\in\mathbb{R}^{c_{h_i}}$ for both RGB and heatmap modalities.

The final re-calibrated feature map can be obtained by taking sigmoid $\sigma$ of the excitation signal and performing channel-wise multiplication with the actual feature map:
\begin{equation}
\mathbf{r'_i}\left(c\right)=2 \cdot\sigma\left(E_{r_i}\left(c\right)\right) \cdot \mathbf{r_i}\left(c\right)
\label{eqn:mmtm_excitation_r}
\end{equation}
\begin{equation}
\mathbf{h'_i}\left(c\right)=2 \cdot\sigma\left(E_{h_i}\left(c\right)\right) \cdot \mathbf{h_i}\left(c\right)
\label{eqn:mmtm_excitation_h}
\end{equation}
We take $dim(\mathbf{Z_i}) = \frac{1}{4}\left(dim\left(\mathbf{S_{r_i}}\right)+dim\left(\mathbf{S_{h_i}}\right)\right)$, as suggested in~\cite{au}. In our case, $dim\left(\mathbf{S_{r_i}}\right) = dim\left(\mathbf{S_{h_i}}\right)$.

To design the complete transfer function $T\left(r,h\right)$, we use this MMTM block for feature reactivation at the last and second last feature maps, as given in Figure~\ref{fig:complete_model}. This is because we want the network to learn some initial representation and then, the higher-level features of both streams can be utilized by each other. At the deeper layers, using this transfer function can help both individual unimodal streams to learn better representation by benefiting from the information extracted by the other stream.

\subsection{CNN Backbone}

We believe that gaze is a relatively low-level feature and thus, a shallow architecture is effective to learn relevant features. So, we designed a custom CNN backbone based on Residual Blocks~\cite{at} as given in Figure~\ref{fig:complete_model}.

It starts with a $3\!\times\!3$ convolution layer followed by Batch Normalization, ReLU and maxpool with kernel size $\left(2,2\right)$ and stride $\left(2,2\right)$. This is followed by $3$ modules, each having $2$ residual blocks followed by a maxpool layer. The intermediate feature output at the end of each module has $64$, $128$ and $256$ channels respectively. Both RGB and heatmap modalities are processed using identical backbones to maintain symmetry, as it creates identical feature dimensions to ease and balance other operations in the network.

\subsection{Aggregation Function}

The aggregation function $A\left(r,h\right)$ is used to fuse the final feature map of both streams in order to get a hybrid representation. This is done by channel-wise concatenation of the individual feature maps, and then passing this through a residual block. The output of this is a hybrid feature map having relevant information from both RGB and heatmap modalities.

\subsection{FLAME}
\label{sec:FLAME}
The complete architecture of FLAME, as depicted in Figure~\ref{fig:complete_model}, is designed from the previously described individual units. The two separate CNN backbones take $120\!\times\!120$ RGB eye patch and $120\!\times\!120\!\times\!28$ eye landmark heatmap. Both learn features independent of each other until the last two modules where both the features are fused into one another by the transfer function $T\left(r,h\right)$, which consists of two MMTM blocks. Thereafter, using the two final feature maps, a hybrid feature map is obtained by the aggregation function $A\left(r,h\right)$. This feature map contains the information of the gaze direction as well as the anatomical information of the eye and is a better descriptor of the gaze direction.

It is then flattened and concatenated with the head pose angles and is passed through two dense layers of 512 neurons each, followed by a two-dimensional output layer for the pitch and yaw gaze angles directly in the World Coordinate System~(WCS). In the dense layers, dropout is used with $p_{dropout} = 0.2$ and ReLU is used as activation function. The output layer, however, is kept as linear with no activation function.
 
\subsection{Loss Function}

This model is end-to-end trainable using the gaze angle ground truth. The 3D angular loss can be defined as
\begin{equation}
\mathcal{L}_{ang}\left(\hat{\boldsymbol{g_p}},\hat{\boldsymbol{g_t}}\right) = \arccos{\frac{\hat{\boldsymbol{g_p}}.\hat{\boldsymbol{g_t}}}{|\hat{\boldsymbol{g_p}}||\hat{\boldsymbol{g_t}}|}},
\label{eqn:3d_angular_loss}
\end{equation}
where $\hat{\boldsymbol{g_p}}$ and $\hat{\boldsymbol{g_t}}$ represent the predicted and true gaze vectors respectively. This loss, given by Equation~\ref{eqn:3d_angular_loss}, directly gives the angular deviation of predicted gaze vector $\hat{\boldsymbol{g_p}}$ from the true gaze vector $\hat{\boldsymbol{g_t}}$ in 3D space.

However, training directly on this loss function, as used by a few models~\cite{ao}, is not ideal because the gradient of this loss, given by
\begin{equation}
\frac{\partial \mathcal{L}_{ang}\left(x\right)}{\partial x} = \frac{1}{\sqrt{1-x^2}} \quad\mathrm{where}\quad x = \frac{\hat{\boldsymbol{g_p}}.\hat{\boldsymbol{g_t}}}{|\hat{\boldsymbol{g_p}}||\hat{\boldsymbol{g_t}}|},
\label{eqn:arccos_derivative}
\end{equation}
increases sharply with decrease in loss and approaches $\infty$ as the loss approaches $0$ ($x$ approaching 1). So, training on this loss makes the network unstable with decrease in loss.

Thus, we define a vector difference loss for training, which is the sum of squared errors of each individual component of the gaze vector $g_k = \left(g_k^x, g_k^y, g_k^z\right)$ where $k=\left\{p,t\right\}$, given by
\begin{equation}
\mathcal{L}_{vec}\left(\hat{\boldsymbol{g_p}},\hat{\boldsymbol{g_t}}\right) = \left(g_p^x\!-\!g_t^x\right)^2\!\!+\!\left(g_p^y\!-\!g_t^y\right)^2\!\!+\!\left(g_p^z\!-\!g_t^z\right)^2.
\label{eqn:vector_loss}
\end{equation}
For evaluation, we use the 3D angular loss itself as per Equation~\ref{eqn:3d_angular_loss}.

\section{Implementation Details}
\label{sec-impl}

\subsection{Datasets}
We perform the experiments with two benchmark datasets: ColumbiaGaze~\cite{bw}, EYEDIAP~\cite{bx}. EYEDIAP is a video dataset with 16 subjects, with different head motion and gaze target settings. We select every $20$-th frame of the HD screen target videos (continuous and discrete) under both static and moving head pose of all subjects for our experiments. ColumbiaGaze is a dataset with discrete values of head pose and targets, having 56 subjects, each with a combination of 7 horizontal and 3 vertical gaze targets. We select all the images from this dataset. For all these images, the ones on which OpenFace outputs could not be obtained were left out of the final dataset. For our experiments, we use the raw images without any re-alignment.

\subsection{Preprocessing}
We first extract face crops of the ColumbiaGaze and EYEDIAP images using RetinaFace~\cite{by}, and zero-pad it to a uniform size of $384\!\times\!480$~(4:5 ratio). As ColumbiaGaze dataset was of higher resolution, we rescale the padded 4:5 crops to this dimension. Then, we run OpenFace~2.0 to obtain eye landmarks, and generate the heatmap for all the data with dimension $d=2$, $\Sigma=\mathbb{I}_{2}$ (2D identity matrix) and $\mu=\left(x_i,y_i\right), i=1,2,\ldots,28$ where $\left(x_i,y_i\right)$ are the obtained eye landmark coordinates, following the method discussed in Section~\ref{sec:heatmap_generation}. We then divide the complete dataset into cross-subject split of 8:1:1 for train, validation and test sets, respectively.

\subsection{Training and Optimization}
\label{sec:settings}

The network is trained with randomly initialized weights for 200 epochs with a batch size of 8, using Adam Optimizer with an initial learning rate of $10^{-4}$, and is decreased by a factor of $0.5$ after $85$-th, $120$-th and $175$-th epoch. During training, $120\!\times\!120$ pixels eye patch of either left or the right eye is picked randomly for each image sample along with the corresponding landmark heatmap of that eye having same resolution, and fed to the network. The eye crop is obtained by calculating the centre of the eye using the four corner landmarks~(8, 14, 11, 17 in Figure~\ref{fig:eye_landmarks}), and taking a $120\!\times\!120$ patch across it with the obtained centre. The same operation is done with landmark heatmap to maintain alignment between both inputs. Head pose is used as available with the datasets, but can also be obtained from facial analysis toolkits like OpenFace~2.0 with sufficient precision. The model is then trained against ground truth gaze angles using the vector difference loss given by Equation~\ref{eqn:vector_loss}.

\subsection{Experimental Settings}
\label{sec:experiments}

In order to conduct an ablation study of the proposed method, we implemented three other models in addition to our main approach of FLAME. All four experimental settings are described below:
\newline
\textbf{FLAME}: Our original approach as proposed in Section~\ref{sec:FLAME} with settings as described in Section~\ref{sec:settings} on all the datasets.
\newline
\textbf{FLAME-aggregation-only~(F-AO)}: The original architecture without the transfer function $T\left(r,h\right)$. This is to evaluate the impact of using multimodal transfer.
\newline
\textbf{FLAME-additive-fusion~(F-AF)}: Another approach to fuse both modalities, where aggregation function is the element-wise addition of the final feature map of both modalities. This is to analyse the effectiveness of aggregation function $A\left(r,h\right)$. There is no $T\left(r,h\right)$ here.
\newline
\textbf{FLAME-baseline~(F-B)}: This is the baseline approach that performs gaze estimation using only the RGB images and proposed CNN backbone. It is to see the contribution of the entire heatmap modality and proposed novelty.

\section{Results}
\label{sec-res}

The performance of our method can be evaluated by Equation~\ref{eqn:3d_angular_loss}, the angular error between true and predicted gaze vectors in 3D space. Results are given in Table~\ref{tab:results} for both cross-subject and cross-dataset validation to see how well the model performs on unencountered subjects~(cross-subject evaluation), and also on completely unknown setup~(distance of subjects, camera settings, etc.) and range of ground truth~(cross-dataset evaluation).

\begin{table}[h!]
\centering
\caption{Results on cross-subject, cross-dataset evaluation.}
\begin{tabular}{|r|c|c|}
\hline
\diagbox{Train}{Test}  & ColumbiaGaze & EYEDIAP \\
\hline
ColumbiaGaze & $\hphantom{1}4.64^{\circ}$ & $12.53^{\circ}$ \\
EYEDIAP & $12.83^{\circ}$ & $\hphantom{1}4.62^{\circ}$ \\
\hline
\end{tabular}
\label{tab:results}
\end{table}

\begin{figure}[htbp]
\centering
\begin{subfigure}[b]{0.23\textwidth}
\includegraphics[width=\textwidth]{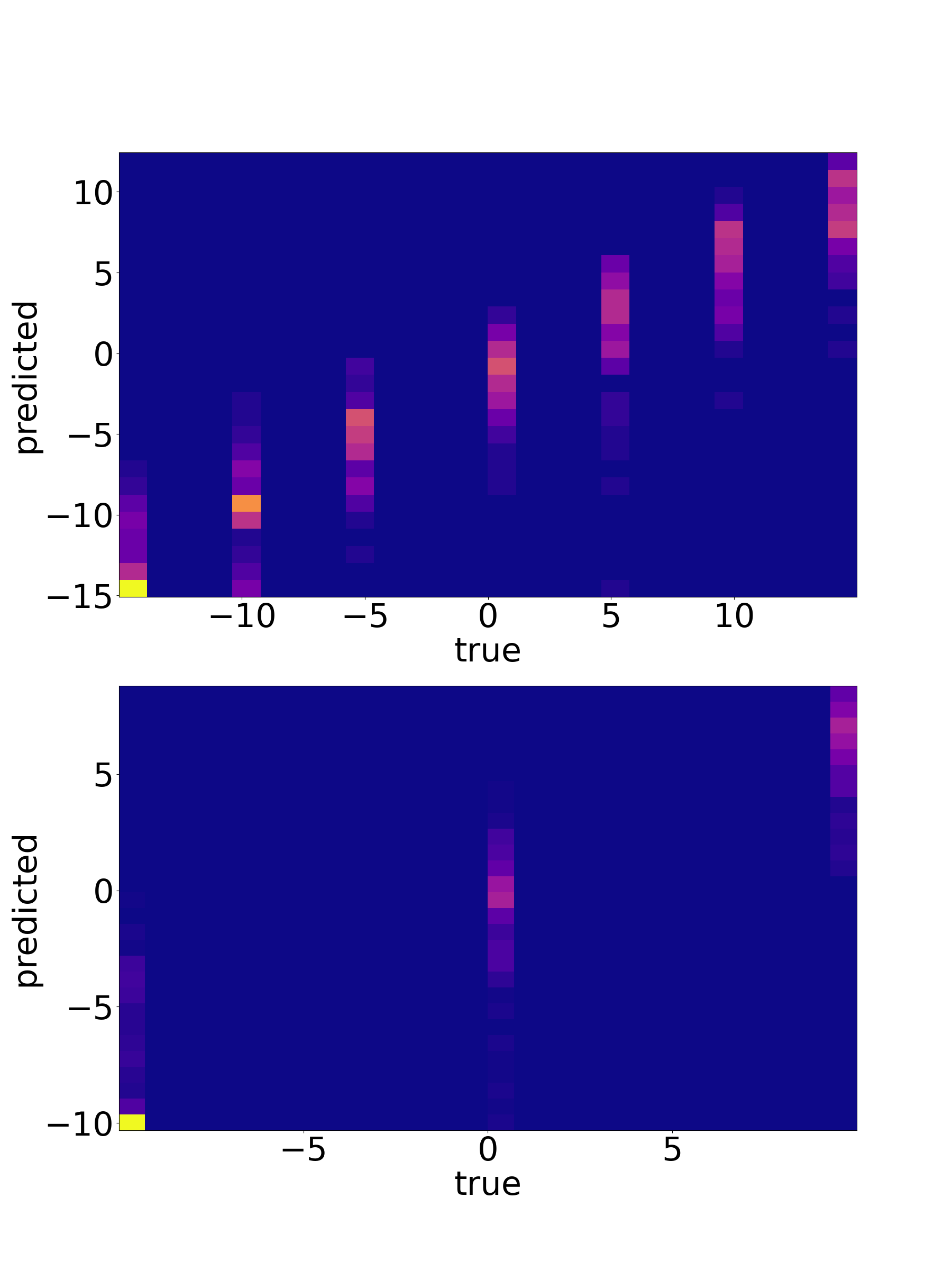}
\caption{ColumbiaGaze}
\label{rfidtest_xaxis}
\end{subfigure}
\begin{subfigure}[b]{0.23\textwidth}
\includegraphics[width=\textwidth]{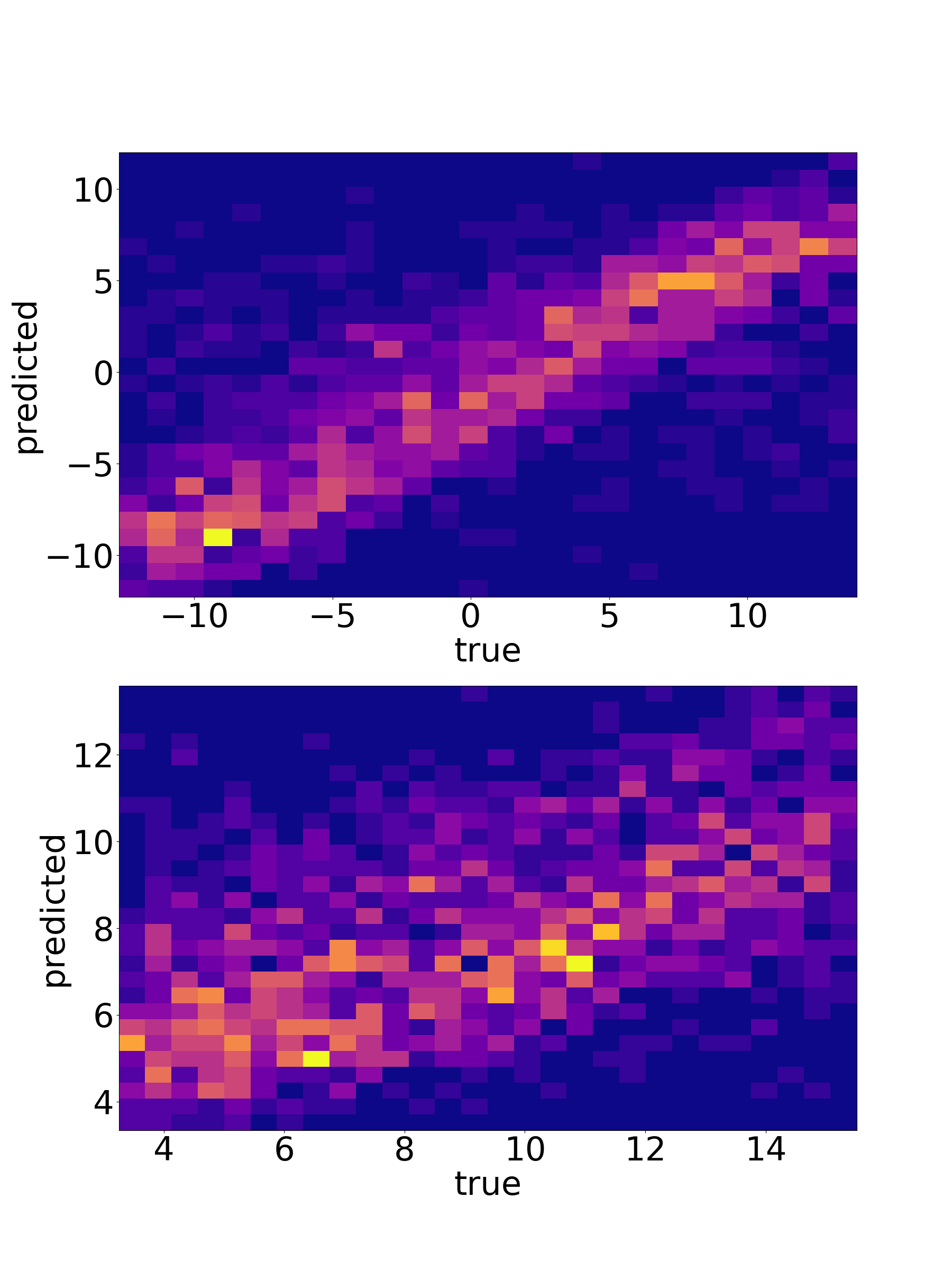}
\caption{EYEDIAP}
\label{rfidtest_yaxis}
\end{subfigure}
\caption{Analysis of predictions versus true labels. Top heatmaps are evaluating the yaw angle and bottom heatmaps are evaluating the pitch angles.}
\label{fig:predictions}
\end{figure}

Figure~\ref{fig:predictions} represents the relation between prediction and truth values for both datasets. They pretty much follow a linear relationship which further proves the correctness of our methodology.

In Table~\ref{tab:ablation_results} we can see that when the model was not provided with any previously encountered subjects, it was able to extract the information from eye landmarks and reduce the mean error by $1.29^{\circ}$ and $0.7^{\circ}$ as compared to RGB only representing the baseline~(F-B). However, to still analyse the factors contributing to errors, it can be understood that we are using 2D facial landmarks, and not 3D. This might fail to capture the precise orientation and shape of the eye in 3D space, and may not describe accurate anatomical structure. Additionally, there can be significant error in eye landmark extraction, particularly when the eye is not fully visible, or when the person is looking directly into the camera. These can be larger for the cornea and pupil landmarks which have finer boundaries.

Cross-dataset error is predominantly because the two datasets are very different to one another in many parameters such as range of gaze angles, distance of subject, or camera parameters. However, although these values are not directly comparable to other methods due to difference in datasets used for cross-dataset evaluation, the performance is competitive to them.

\subsection{Ablation Study}

It is also important to understand the contribution of all the components proposed in our methodology. For this, we implemented three other architectures as described in Section~\ref{sec:experiments}. The results obtained are given in Table \ref{tab:ablation_results}.

\begin{table}[h!]
\centering
\caption{Mean and standard deviation (std) of angular error for different experimental settings}
\begin{tabular}{c|c|c||c|c}
\hline
Experiment Setup & \multicolumn{2}{c||}{ColumbiaGaze} & \multicolumn{2}{c}{EYEDIAP} \\
\hline
& mean & std & mean & std \\
\hline \hline
F-B & $5.93^{\circ}$ & $3.20^{\circ}$ & $5.32^{\circ}$ & $3.08^{\circ}$ \\
F-AF & $5.88^{\circ}$ & $3.06^{\circ}$ & $5.30^{\circ}$ & $3.03^{\circ}$ \\
F-AO & $5.06^{\circ}$ & $3.13^{\circ}$ & $4.80^{\circ}$ & $3.02^{\circ}$ \\
\hline
\textbf{FLAME} & \textbf{4.64}$^{\circ}$ & \textbf{2.86}$^{\circ}$ &  \textbf{4.62}$^{\circ}$ & \textbf{2.93}$^{\circ}$ \\
\hline
\end{tabular}
\label{tab:ablation_results}
\end{table}

\begin{figure}[htbp]
\centering
\includegraphics[width=\linewidth]{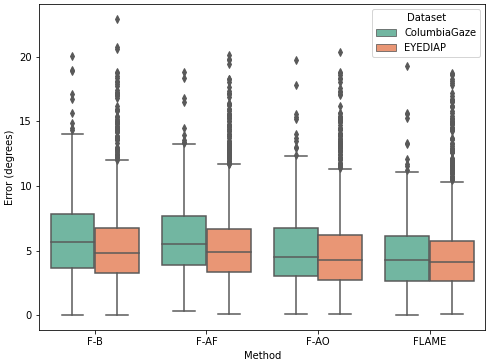}
\caption{Distribution of errors of all experimental settings}
\label{fig:error_distribution}
\end{figure}

We can see that FLAME outperforms all other experimental settings. Figure~\ref{fig:error_distribution} depicts the distribution of error for EYEDIAP and ColumbiaGaze datasets on all four experimental settings through a box plot.

There is a difference in mean error of $0.18^{\circ}$ for EYEDIAP and $0.42^{\circ}$ for ColumbiaGaze between F-AO and FLAME, where the absence of transfer function was the only difference. This shows its contribution.

Further, F-AO and F-AF differ in aggregation function $A\left(r,h\right)$. We compare our approach with element-wise addition method in F-AF, where $A\left(r,h\right)=\mathbf{r_4}+\mathbf{h_4}$. The error difference highlights the effectiveness of our aggregation function for creating a hybrid feature map.

At last, the contribution of our complete methodology can be seen through the difference of FLAME with F-B.

We also compared the performance of FLAME with different input resolutions. In practical settings, having high-resolution eye images is a constraint due to limited camera resolution, distance of subject from it, etc. On ColumbiaGaze, we obtained error of $4.79^{\circ}$ in $60\!\times\!60$ pixels and $5.5^{\circ}$ on $30\!\times\!30$ pixels image resolution. Based on this, we would propose $120\!\times\!120$ pixels as optimum and $60\!\times\!60$ pixels as the minimum threshold of input resolution with FLAME.

\subsection{Comparison with State-of-the-art}

We also compare our method with various state-of-the-art. The results are given in Table~\ref{tab:comparison_SOTA}. All of them are for leave-one-subject-out settings (or cross-subject evaluation).

\begin{table}[h!]
\centering
\caption{Mean 3D angular error as reported by different SOTA methods: person-specific methods (top), person-independent methods (bottom)}
\begin{tabular}{c|c|c}
\hline
Method&ColumbiaGaze & EYEDIAP \\
\hline \hline
Yu \textit{et al.} (sup. HCS)~\cite{bl} & $5.25^{\circ}$ & $7.09^{\circ}$ \\
Yu \textit{et al.} (sup. WCS)~\cite{bl} & {3.54}$^{\circ}$ &  $6.79^{\circ}$ \\
Yu \textit{et al.} (unsup.)~\cite{bl} & $7.15^{\circ}$ & $8.2^{\circ}$  \\
\hline
Zhang \textit{et al.}~\cite{ba} & - & $6.0^{\circ}$  \\
Zhang \textit{et al.}~\cite{am}\textsuperscript{\#}& - & $7.37^{\circ}$ \\
Zhang \textit{et al.}~\cite{an}\textsuperscript{\#}& - & $6.79^{\circ}$ \\
Cheng \textit{et al.}~\cite{bz}& - & $5.3^{\circ}$ \\
Cheng \textit{et al.}~\cite{bm}& - & $5.17^{\circ}$ \\
Fischer \textit{et al.}~\cite{ak}\textsuperscript{*}& - & $6.4^{\circ}$\\
Chen \textit{et al.}~\cite{ca}\textsuperscript{*}& - & $5.9^{\circ}$\\
Kellnhofer \textit{et al.}~\cite{cb}\textsuperscript{\#}& - & $5.36^{\circ}$\\
Yu \textit{et al.}~\cite{bc}& - & $8.54^{\circ}$\\
\textbf{FLAME (ours)} & \textbf{4.64}$^{\circ}$ & \textbf{4.62}$^{\circ}$ \\
\hline
\multicolumn{3}{c}{*\footnotesize{error metric as reported by~\cite{bz}}, \textsuperscript{\#}\footnotesize{error metric as reported by~\cite{bm}}}
\end{tabular}
\label{tab:comparison_SOTA}
\end{table}

It can be seen that our method performs significantly better than all of the person-independent methods for gaze estimation. This shows the impact of using eye landmark heatmaps to utilize anatomical information of the eye along with the RGB image. When comparing to the models of~\cite{bl}, a person-specific method (based on few-shot calibration), our approach performs better than all 3 methods on both datasets, except for one model in ColumbiaGaze. A larger error than person-specific models can be because when the model gets the test samples directly, it can extract additional information like orientation in 3D space, PERCLOS (\% of eye closed), and other salient anatomical features, which are not possible to capture using 2D eye landmarks. For ColumbiaGaze, SOTA benchmarks are not available for many models.

Thus, with all these analyses and experimentation, we can justify that our approach holds key to a precise gaze estimation.

\section{Conclusion}
\label{sec-con}

We presented an approach of using 2D eye landmarks along with RGB images to perform 3D gaze estimation by incorporating anatomical features of the eye without any person-specific calibration. In our approach, we obtained a 2D Gaussian probability distribution heatmap of eye landmarks from the RGB image itself and then used a two-stream CNN-based network to effectively extract relevant features. We used this along with head pose to predict gaze angles. Our method gives better results than all state-of-the-art methods on two benchmark datasets. This proves that eye landmarks can play a vital role in incorporating anatomical information to predict gaze more accurately.

We believe more precise systems can be designed if we can incorporate more information related to the anatomical aspects of the eye, and use eye landmarks in other ways. As of now, we leave that for further study.


{\small
\bibliographystyle{ieee}
\bibliography{egpaper}
}

\end{document}